\documentclass{article}
\usepackage{subcaption}
\usepackage[superscript]{cite}  
\usepackage{rotating}
\usepackage{graphicx}
\usepackage{longtable}
\usepackage{pdflscape}
\usepackage[hyphens]{url}
\usepackage{pdflscape} 
\usepackage{enumitem}

\usepackage{hyperref}
\usepackage{booktabs}
\usepackage[margin=1in]{geometry}
\usepackage{tabularx} 

\hypersetup{breaklinks=true}

\usepackage{soul}

\usepackage{authblk}  
\hypersetup{colorlinks=true, urlcolor=blue, linkcolor=blue, citecolor=blue}  

\providecommand{\fnm}[1]{#1}
\providecommand{\sur}[1]{#1}
\providecommand{\orgdiv}[1]{#1}
\providecommand{\orgname}[1]{#1}
\providecommand{\orgaddress}[1]{#1}
\providecommand{\city}[1]{#1}
\providecommand{\country}[1]{#1}

\providecommand{\email}[1]{\texttt{#1}}

\title{Data-Driven Analysis of AI in Medical Device Software in China: Trends of Deep Learning and Traditional AI Based on Regulatory Data}

\author[1,*]{\fnm{Yu} \sur{Han}\thanks{These authors contributed equally to this work.}} 
\author[2,*]{\fnm{Aaron} \sur{Ceross}\thanks{These authors contributed equally to this work.}} 
\author[3]{\fnm{Sarim} \sur{Ather}} 
\author[1,4]{\fnm{Jeroen H.M.} \sur{Bergmann}} 

\affil[1]{\orgdiv{Institute of Biomedical Engineering, Department of Engineering Science}, \orgname{University of Oxford}, \city{Oxford},\country{United Kingdom}}
\affil[2]{\orgdiv{Birmingham Law School}, \orgname{University of Birmingham}, \orgaddress{\city{Birmingham}, \country{United Kingdom}}}
\affil[3]{\orgdiv{NHS Foundation Trust}, \orgname{University of Oxford}, \country{United Kingdom}}
\affil[4]{\orgdiv{Department of Technology and Innovation}, \orgname{University of Southern Denmark}, \city{Odense}, \country{Denmark}}

\affil[*]{\email{yu.han@eng.ox.ac.uk}, \email{a.w.k.ceross@bham.ac.uk}}
\affil[ ]{\email{sarim.ather@ouh.nhs.uk}, \email{jeroen.bergmann@eng.ox.ac.uk}}

\date{\today} 

\begin{document}
\setcounter{page}{1}
\pagestyle{plain} 
\pagenumbering{arabic} 

\maketitle
\abstract{
Artificial intelligence (AI) in medical device software (MDSW) represents a transformative clinical technology, attracting increasing attention within both the medical community and the regulators. In this study, we leverage a data-driven approach to automatically extract and analyze AI-enabled medical devices (AIMD) from the National Medical Products Administration (NMPA) regulatory database. The continued increase in publicly available regulatory data requires scalable methods for analysis. Automation of regulatory information screening is essential to create reproducible insights that can be quickly updated in an ever changing medical device landscape. More than 4 million entries were assessed, identifying 2,149 MDSW registrations, including 385 standalone applications and 1,764 integrated within medical devices, of which 43 were AI-enabled. It was shown that the leading medical specialties utilizing AIMD include respiratory (20.5\%), ophthalmology/endocrinology (12.8\%), and orthopedics (10.3\%). This approach improves the speed of data extraction, enabling more timely comparisons—though careful validation remains essential when interpreting these results. This study provides the first extensive, data-driven exploration of AIMD in China, showcasing the potential of automated regulatory data analysis in understanding and advancing the landscape of AI in medical technology.}


\providecommand{\keywords}[1]{\textbf{\textit{Keywords---}} #1}



\maketitle

\section{China's New Generation Artificial Intelligence Development Plan and Effect to Medical Devices}
In 2017, China’s State Council released its \textit{New Generation Artificial Intelligence Development Plan}, which set forth a strategic roadmap aimed at positioning the country as a global leader in artificial intelligence (AI) technologies~\cite{NewGenerationPlan}. Within the medical devices sector, China has emerged as a significant global producer and consumer, leveraging its large-scale manufacturing capabilities and distribution networks~\cite{morrison2013china}. Given the country’s population of 1.41 billion in 2023, the demand for advanced medical technology is substantial~\cite{chinapopu}. The Chinese medical device market, valued at ¥629 billion RMB (approximately \$88.7 billion USD) in 2019, more than doubled from ¥308 billion RMB in 2015~\cite{deloitte2021indusstry}. China's long-term strategy emphasizes the integration of AI and machine learning into medical domain, with the goal of enhancing clinical support through improved diagnostics and intervention management~\cite{he2019practical}. In line with this vision, the National Medical Products Administration (NMPA), China’s regulatory authority for medical products, has enacted several regulations to support the development and oversight of AI-driven medical devices. Table \ref{tab:regulations} summarizes key regulatory documents issued by the NMPA in relation to AI-based medical device software~\cite{han2024regulatory}. The NMPA’s first relevant guidance was issued in 2015, with a more comprehensive regulatory framework established in 2022. These regulations reference a range of national standards, including those governing software risk classification (YY/T 0664-2008), software engineering practices (GB/T 19003-2008), and medical device quality management (YY/T 0287-2003) \cite{han2024regulatory}.

\begin{table*}[t!]
    \centering
   \begin{tabular}{c|p{7cm}}
   \hline
         \textbf{Date of Publication} & \textbf{Regulatory Document}  \\ \hline
         08/2015 & Guidelines of medical device software registration and review \cite{no50} \\ \hline
         07/2019 & Key Points of Deep Learning Decision-making Assisting Medical Device Software review \cite{keypoints} \\ \hline
         07/2021 & Guidelines for the classification and designation of artificial intelligence medical software \cite{NO47} \\ \hline
         03/2022 & Guidelines of medical device software registration and review\cite{no9}\\ \hline
         08/2022 & Guidelines for the classification and designation of artificial intelligence medical software \cite{no8} \\ \hline
    \end{tabular}
    \caption{Chinese regulatory documents for medical device software}
    \label{tab:regulations}
\end{table*}

According to International Medical Device Regulators Forum (IMDRF), a medical device is defined by its intended use—such as diagnosis, monitoring, treatment, or prevention of disease, injury, or disability; supporting physiological functions; or controlling conception~\cite{aronson2020medical}. For instance, an app that tracks fitness for general wellness is not a medical device, whereas an app designed to support contraception or fertility treatment is. In this study, we focus on Medical Device Software (MDSW), which includes both Software as a Medical Device (SaMD) and Software in a Medical Device (SiMD), provided they meet the IMDRF's definition. These systems support clinical decisions and may be used for purposes such as diagnostic assistance, disease monitoring, image analysis, or treatment planning. Examples include software for radiological image interpretation~\cite{wang2017hospital}, oncology tools for genetic risk prediction~\cite{patel2018enhancing}, ophthalmology solutions for diabetic retinopathy screening~\cite{ting2019artificial}, and general clinical decision support systems~\cite{szolovits2019artificial}.

While prior research has explored the deployment of AI-enabled medical devices (AIMD) in various regulatory landscapes, including notable work on regulatory frameworks and device categorization methods~\cite{liu2024regulatory}, our study introduces an innovative, fully automated approach. By leveraging advanced data science techniques, our methodology filters targeted devices directly from the NMPA regulatory database. This automated system enables rapid, comprehensive identification and analysis of AIMD in under one minute after establishing selection criteria, significantly enhancing the efficiency and precision of regulatory data analysis. This automated approach represents a substantial advancement over previous methods, which often involved manual data collection or semi-automated processes, making it a uniquely scalable and efficient solution for regulatory assessment.

\subsection*{China Software Medical Device Registration Process}

\begin{figure}[!t]
    \centering
    \includegraphics[width=\textwidth]{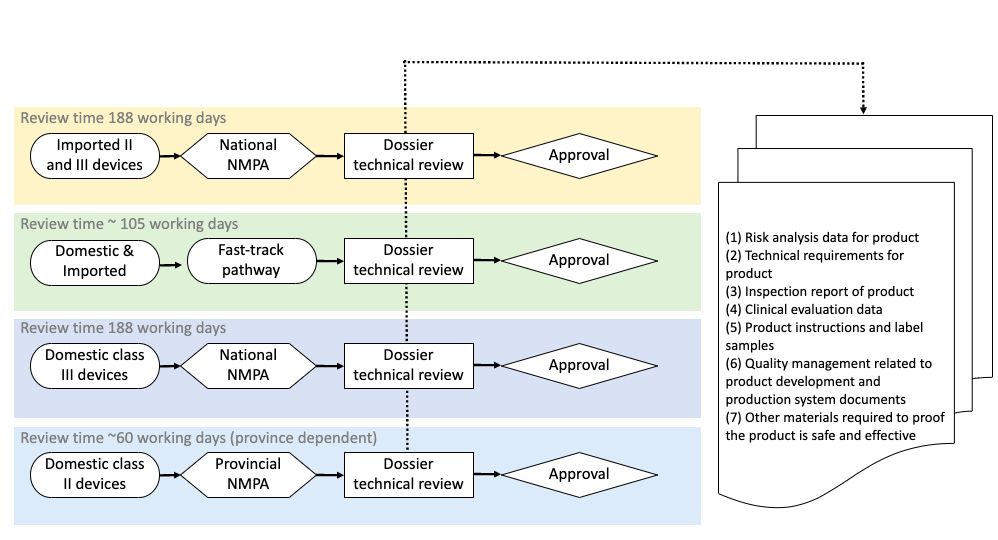}
    \caption{Medical device registration pathways set out by the NMPA. Imported and domestic devices relates to all class II and III devices, for Class I devices, only record-filing is required.}
    \label{fig:process}
\end{figure}

Figure~\ref{fig:process} provides an overview of the medical device registration process administered by the NMPA for various types of devices. The NMPA has also established expedited pathways, including the Innovation Approval~\cite{no83}, Priority Review~\cite{no168}, and Emergency Approval~\cite{no157} processes. The final version of the “Innovation Approval Procedure for Medical Devices,” published on 5 November 2018, outlines three criteria for fast-track approval: a) The applicant must have completed preliminary research on the product and developed a finalized prototype with comprehensive and traceable research data. b) The primary technical mechanism of the product must be novel, inventive, and hold significant clinical value, meaning that the product's performance or safety offers distinct advantages over existing alternatives. c) The applicant must either hold the patent for the core technology of the product in China or have obtained the right to use the patent in China through technology transfer~\cite{no83}. Both domestic and foreign manufacturers are subject to the same application process for innovative device registration.

\subsection*{Unique Device Identification System and Database Background}

The Unique Device Identification (UDI) system is a framework proposed by IMDRF and its intention is to adequately identify medical devices from manufacturing through distribution and patient use. Since joining the IMDRF in 2013 ~\cite{MinYue2019imdrf}, China has been adopting and referencing international regulatory methods when formulating China regulations. China officially adopted the UDI System in 2019, as outlined by two key documents: (i) Pilot Project Plan for UDI System for Medical Devices (No.56, 2019)~\cite{no56} and (ii) Rules for Unique Device Identification System (No. 66, 2019)~\cite{no66}. The NMPA is implementing the adoption of UDI in phases. The strategy is to start with devices that carry the highest risk (Class III) and then proceed to add lower risk devices in each new phase. In January 2021, nine categories of high-risk Class III medical devices were included in the first batch. From June 2022, the second batch of medical devices was included. As of August 2024, there are 4,045,039 products registered with UDI system, with most  being general medical devices (96.34\%), whilst in vitro medical devices (3.66\%) are only making up a small amount of the total registrations\cite{udidata}. The NMPA's UDI database is publicly accessible, allowing patients, healthcare providers, and regulatory authorities to quickly access information about a particular medical device. 

\begin{figure}[t]
    \centering
    \includegraphics[scale=0.18]{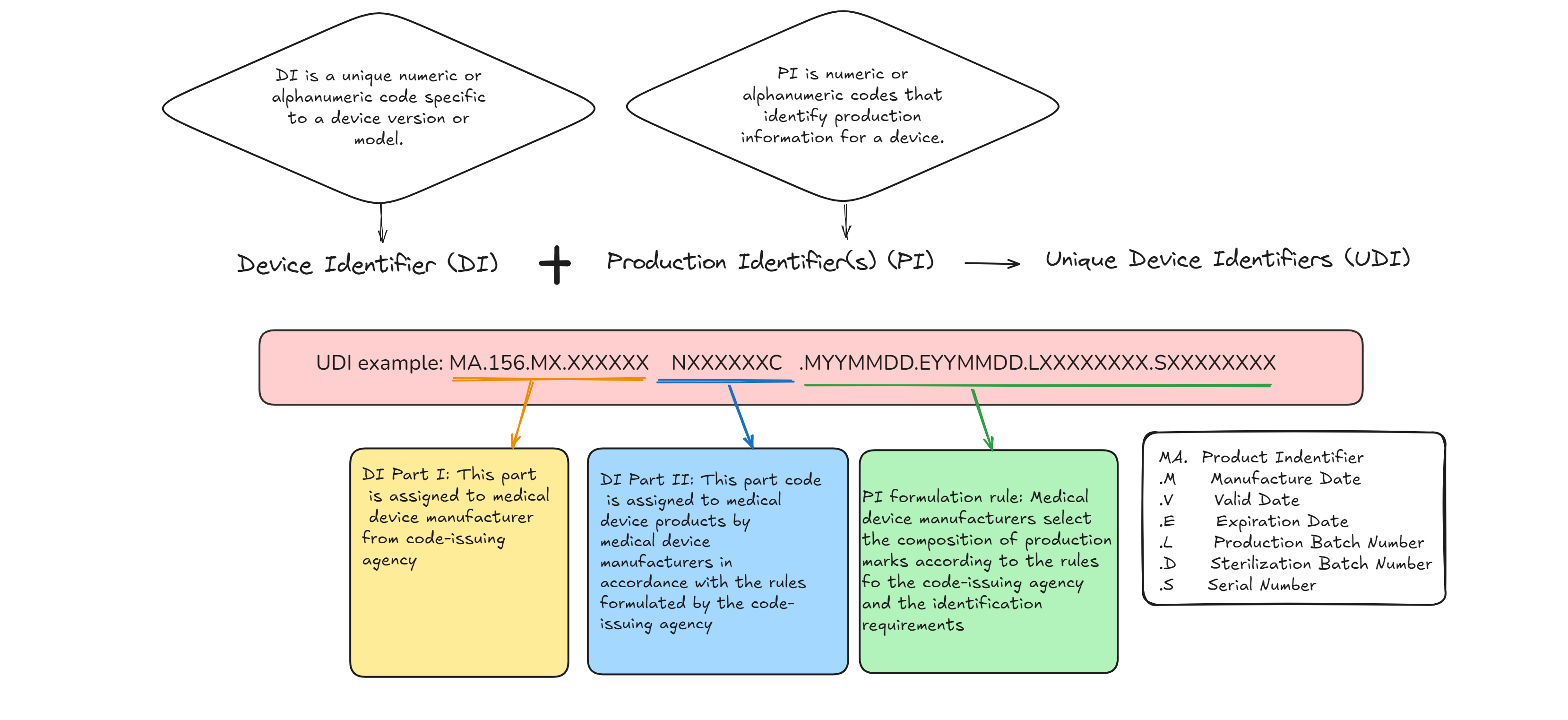}
    \caption{UDI code layout according to YY/T 1630-2018.}
    \label{fig:udi}
\end{figure}

Figure~\ref{fig:udi} illustrates the meaning of the number codes in China's UDI system~\cite{yyt}. A UDI code typically consists of a Device Identifier (UDI-DI) and a Production Identifier (UDI-PI). The UDI-DI refers to a unique numeric or alphanumeric code that is specific to a particular model of medical device and is also used as an access key to information stored in the database. The UDI-PI, on the other hand, identifies the unit of device production with a numeric or alphanumeric code. In China, the UDI-DI is composed of two parts: part I is assigned by the code-issuing agency, while part II is issued by the manufacturer. 

\subsection*{Research Aim and Contribution}
This study aims to evaluate how real-world regulatory database can be used to generate insights into the landscape, classification, and governance of AI-enabled medical devices. While we introduce a rule-based identification method to support this analysis, our core objective is to apply data science in the service of regulatory science.

We are particularly interested in exploring how the Chinese regulatory authority (NMPA) classifies AI-enabled medical devices, and whether there are systematic patterns in device classification that correlate with technical and regulatory factors such as device type (SaMD vs. SiMD), algorithm category (e.g., deep learning vs. traditional AI), clinical function (e.g., diagnosis, monitoring), and product origin (domestic vs. imported). To investigate this, we develop and apply a reproducible, rule-based method that integrates regulatory logic, product classification structures, and naming conventions. This method allows for efficient extraction, identification, and the analysis of AI-related submissions contained within a comprehensive national-level medical device registry.

To guide this investigation, we pose the following primary research question:

\begin{quote}
\textit{What can publicly available regulatory data reveal about the development, deployment, and oversight of AI-enabled medical devices in China?}
\end{quote}

This question is further explored through the following sub-questions:

\begin{quote}
\begin{itemize}
    \item \textit{What types of medical device software (MDSW), including SaMD and SiMD, have been registered in China that involve AI components?}
    \item \textit{Which domestic and international manufacturers are contributing to this domain?}
    \item \textit{What kinds of AI techniques (e.g., deep learning, rule-based systems) are being used, and in what medical specialties or disease areas?}
    \item \textit{How do observed trends correspond with regulatory policy changes, such as updated guidelines or new classifications?}
    \item \textit{Explore how different AI algorithm types are currently represented in regulatory classifications, and whether observable patterns emerge?}
    \item \textit{What methodological challenges arise when using registry data to analyze emerging AI-enabled devices?}
\end{itemize}
\end{quote}

Our approach is rooted in a scalable exploratory data analysis (EDA) framework. We use the rule-based algorithm to classify AI-enabled medical devices, including SaMD and SiMD, and generate a set of structured observations from the UDI registry. This enables rapid identification of relevant entries at scale and supports systematic monitoring of evolving AI adoption patterns. However, the tool is a means rather than an end. Our central contribution lies in the use of this automated pipeline to empirically examine regulatory practice, track industrial behavior, and understand the emerging medical AI ecosystem through the lens of real-world registry data. In addition to the core objective of evaluating classification patterns using automated identification methods, we also include an exploratory analysis of algorithm types and observed regulatory variations, to support future hypothesis generation. These findings are descriptive only and are not intended to make normative or causal claims.

Our contributions are fourfold:
\begin{enumerate} \item We develop and validate a scalable, reproducible method for extracting AI-related entries from a national regulatory database, providing a foundation for systematic surveillance of AI-enabled medical technologies.
\item We analyze descriptive patterns in regulatory classification, manufacturer distribution, algorithmic design, and clinical domains, offering empirical insights into how AI technologies are integrated into healthcare products.
\item We propose a conceptual framework for leveraging regulatory data to study oversight behavior, highlighting the potential for empirical approaches to inform broader digital health governance.
\item We demonstrate how real-world regulatory data can be used to infer the actual deployment of AI techniques within specific medical device contexts. This is particularly valuable given the limited transparency and explainability of many AI systems, and may inform future policy development and standards-setting.
\end{enumerate}

\section{Methods}
\label{sec:methods}
\subsection*{Definition of MDSW, SiMD, SaMD and AI software}
Medical Device Software (MDSW) has seen widespread adoption in China and encompasses both Software as a Medical Device (SaMD) and Software in a Medical Device (SiMD). The International Medical Device Regulators Forum (IMDRF) defines SaMD as software intended for one or more medical purposes, capable of performing these functions without being part of a hardware medical device~\cite{samd}. Medical purposes, are designed for a range of clinical applications, such as disease diagnosis, prevention, monitoring, treatment, and physiological support, as well as life-sustaining functions and reproductive health management \cite{IMDRF_SaMD}. In China, SaMD can be identified by a six-digit classification code beginning with the number 21, as outlined in the medical device catalogue~\cite{catalog}, which is provided in Appendix 1. SaMD functions independently as medical software, while Software-inclusive Medical Devices (SiMD) involve additional components, such as hardware, with the software and hardware working together.


According to NMPA Regulation Guiding Principles for Registration Review of Artificial Intelligence Medical Devices\cite{NMPANo47}, an AI medical device referred to  medical device that utilises artificial intelligence technology to achieve its intended use (i.e., medical purposes) by using "medical device data." Medical device data refers to objective data generated by medical devices for medical purposes. This includes medical imaging data generated by medical imaging devices (such as X-ray, CT, MRI, ultrasound, endoscope, optical imaging and so on), physiological parameter data generated by the medical devices (such as ECG, EEG, blood pressure and so on), and in vitro diagnostic data (such as pathological images, microscopic images, invasive blood glucose waveform data and so on). Therefore, an AI medical device refers to a medical device that utilises artificial intelligence technology to process and analyse medical (device) data for medical purposes.

\subsection*{Pre-processing}
\subsubsection*{Description of database}
All data analysis was conducted using RStudio (version 2022.12.0, Posit Software, PBC). The exploratory data analysis was performed on the \nolinkurl{UDID_FULL_RELEASE_20240801.zip} dataset obtained from the NMPA website~\cite{daa}. The dataset contains information on medical devices registered with the NMPA UDI system from 2016 to 2024, consisting of approximately 4 million products, each of product labelled with 48 variables, such as product description, version number, generic name, specification, product number, device classification and so on.

\begin{figure}[t!]
    \centering
    \includegraphics[scale=0.65]{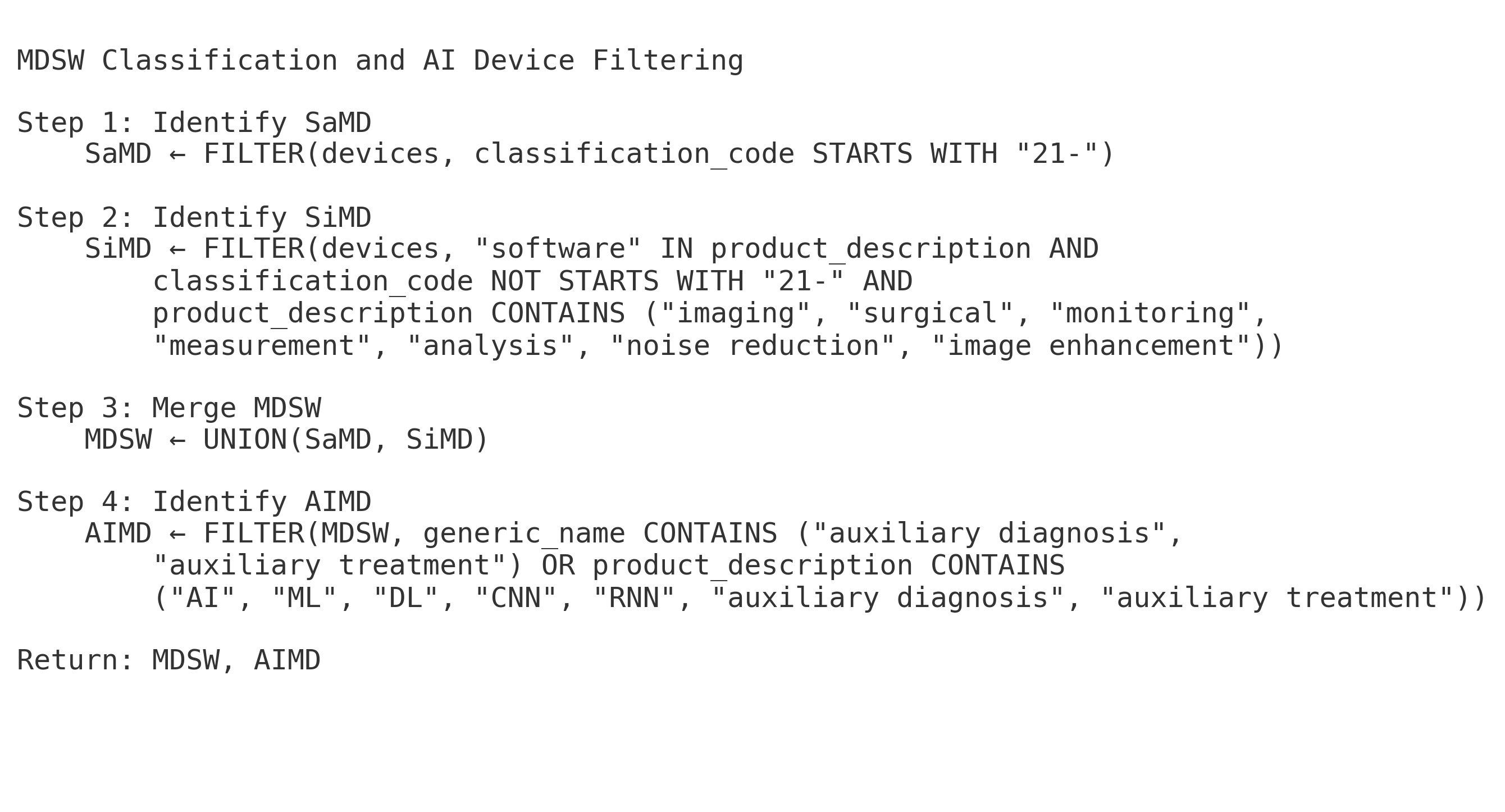}
    \caption{Pseudocode for MDSW and AIMD Identification Process}
    \label{fig:pse}
\end{figure}

\subsection*{Search Strategy}
\begin{figure}[t!]
    \centering
    \includegraphics[scale=0.45]{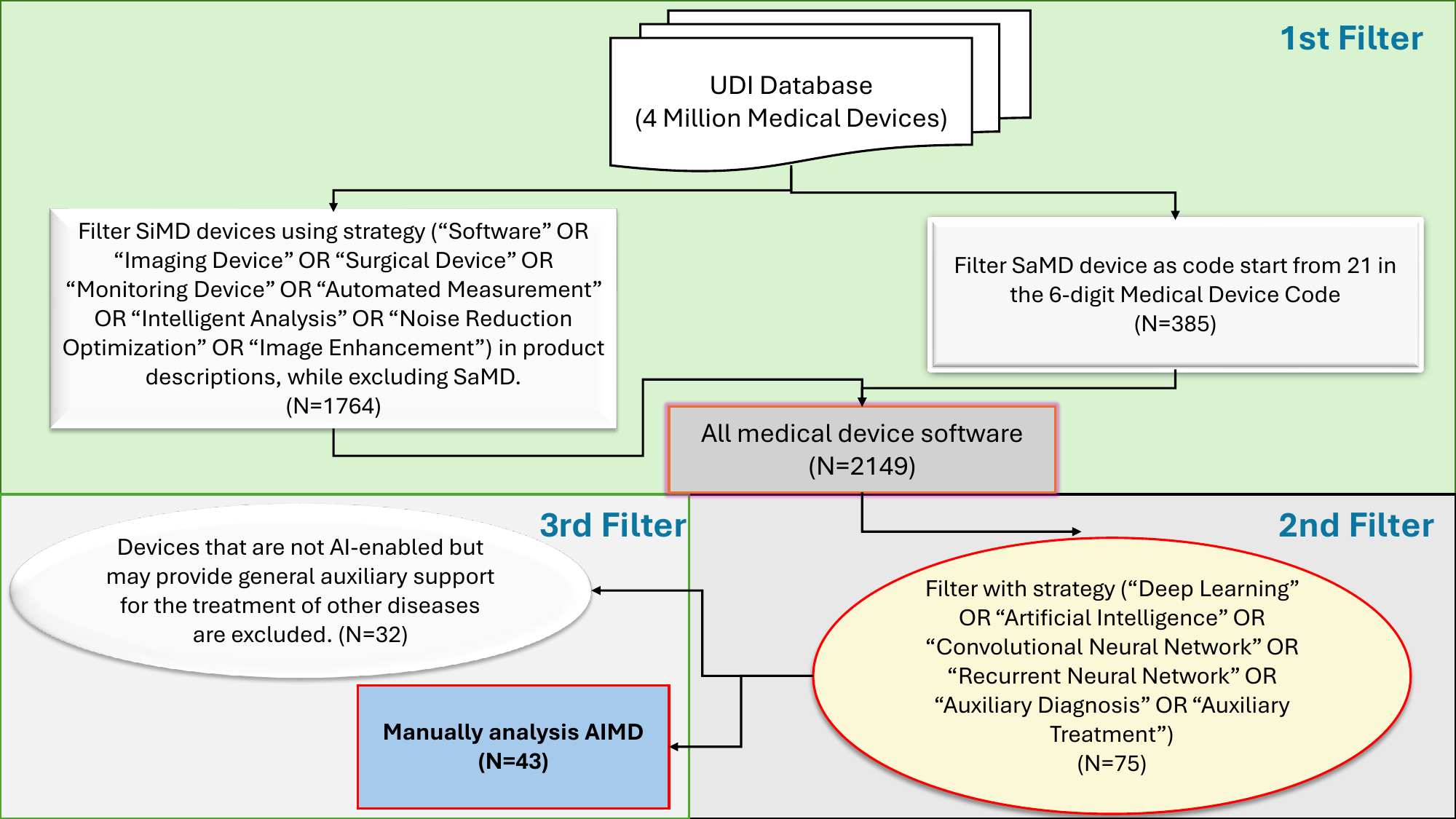}
    \caption{Flowchart for the selection process for the identification of software and AIMD}
    \label{fig:filter process}
\end{figure}

The data was filter using a two-layers strategy ~\cite{nmpa-udi}, as shown in pseudo code in Figure \ref{fig:pse} and flowchart in Figure~\ref{fig:filter process}. The first layer consisted of filtering out MDSW, which consist of filter out SiMD and SaMD. whilst a second layer was applied to then identify AIMD from first layer result, using keywords shown in Figure \ref{fig:filter process}.  After generating the final list, the authors manually filtered it again to remove any devices that did not fall in the AI domain, for example, the devices are not AI-enabled by may provide auxiliary support for the treatment of other diseases.   

The filtering process for identifying AI-enabled medical devices (AIMD) was conducted in a structured, multi-stage approach to systematically refine the dataset from the UDI database, which contains approximately 4 million medical devices. First, medical device software (MDSW) was extracted by isolating both Software as a Medical Device (SaMD) and Software in a Medical Device (SiMD). SaMD was identified using the six-digit classification code rule, where codes beginning with \textbf{“21-”} denote standalone software-based medical devices. SiMD, which lacks a specific classification code, was identified using a keyword-based filtering approach applied to product descriptions, ensuring relevance to medical applications. The keywords we adopted included \textbf{(“software” OR “imaging device” OR “surgical device” OR “monitoring device” OR “automated measurement” OR “intelligent analysis” OR “noise reduction optimization” OR “image enhancement"}.Devices already categorized as SaMD (21-code rule) were explicitly excluded to prevent misclassification. This process resulted in a total of 2,149 MDSW devices.

To further identify AI-enabled medical devices. According to the Chinese regulation \emph{Guiding principles for naming of medical software}~\cite{naming}, devices which incorporate artificial intelligence must include in their name either the terms (i) Auxiliary diagnosis or (ii) Auxiliary treatment. For example, as indicated in the NMPA regulation, if the radiotherapy contouring software uses non-artificial intelligence technology, it is named "radiotherapy contouring software"; if it uses artificial intelligence technology, it is named "radiotherapy contouring auxiliary decision-making software". So we adopt keywords as searching strategy to filter devices based on the results from first step:  
an additional filtering layer was applied to the MDSW dataset using strategy: \textbf{(“Artificial Intelligence” OR “Machine learning” OR “Deep learning” OR “Reinforcement learning” OR “Auxiliary diagnosis” OR “Auxiliary treatment” OR “Convolutional Neural Network (CNN)” OR “Recurrent Neural Network (RNN)}. This automated step initially identified 75 candidate AIMD. However, due to regulatory wording, certain false positives emerged, particularly among devices labeled with “Auxiliary Diagnosis” or “Auxiliary Treatment,” which, in some cases, referred to general software support rather than AI-driven functionalities. To refine the dataset, a manual verification step was conducted, eliminating 32 devices that did not exhibit AI functionality, resulting in a final count of 43 AI-enabled medical devices. This structured filtering approach ensured both high precision and recall, effectively distinguishing genuinely AI-enabled medical devices from broader medical software applications.

 \section{Results}
\label{sec:results}

\subsection{Devices amount and category}

\begin{figure}[t!]
    \centering
    \includegraphics[scale=0.6]{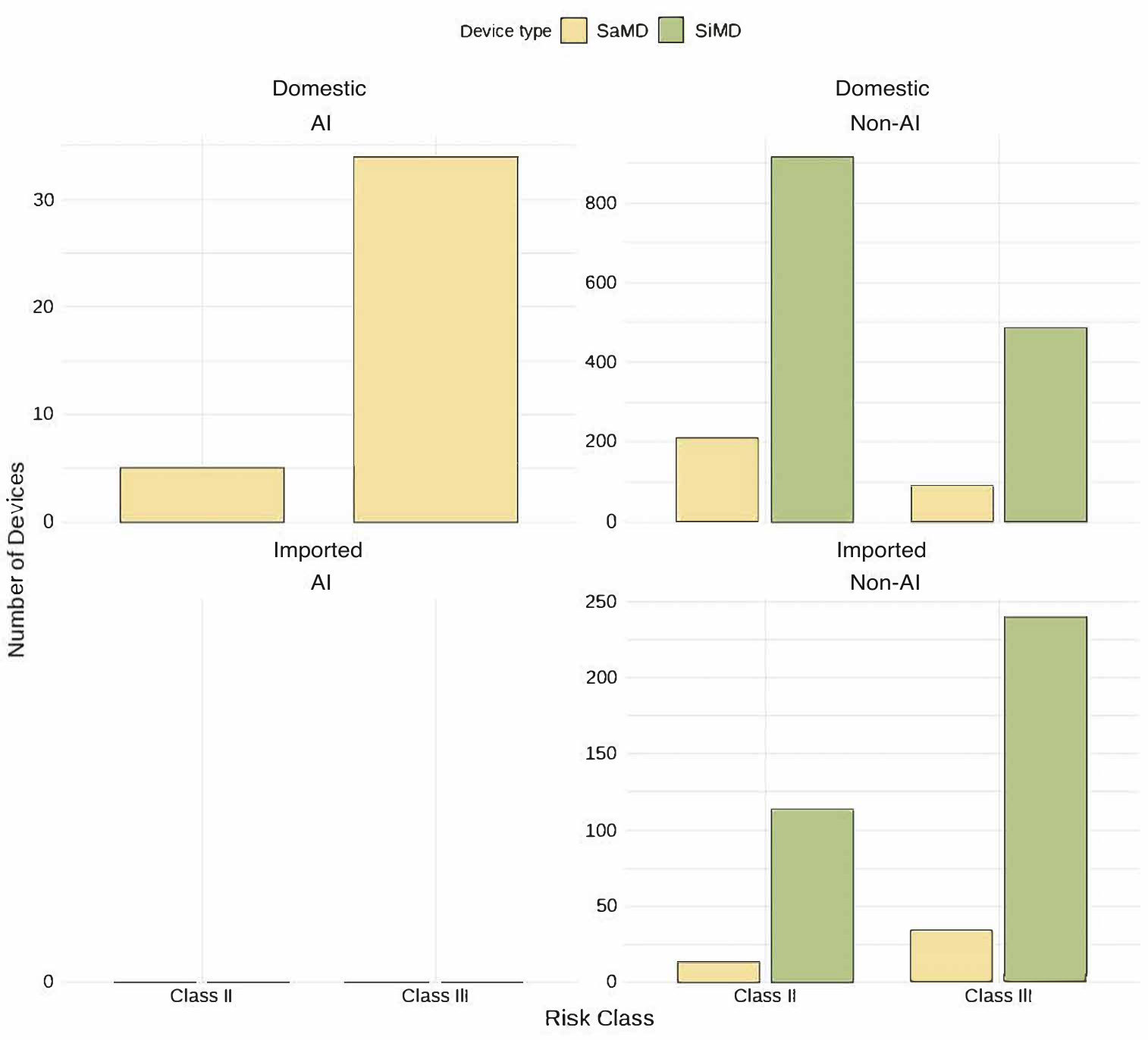}
\caption{Domestic and Imported SaMD \& SiMD Devices: AI vs. Non-AI Distribution. Different scales are applied to each subplot to better illustrate variations across categories.}

    \label{fig:bar}
\end{figure}

A total of 2,149 MDSW were identified, with 385 classified as SaMD and 1,764 classified as SiMD devices, as shown in Figure \ref{fig:bar} . The majority of devices are domestic products, with the largest category being domestic SiMD Non-AI devices, comprising over 60\% of the total of whole medical software. Imported SiMD non-AI devices constitute approximately 11\% of the total. AI-enabled devices are predominantly concentrated in domestic SaMD, making up more than 90\% of all AI devices identified. Notably, no SiMD device was found to incorporate AI.

\begin{figure}[!t]
\includegraphics[width=\textwidth]{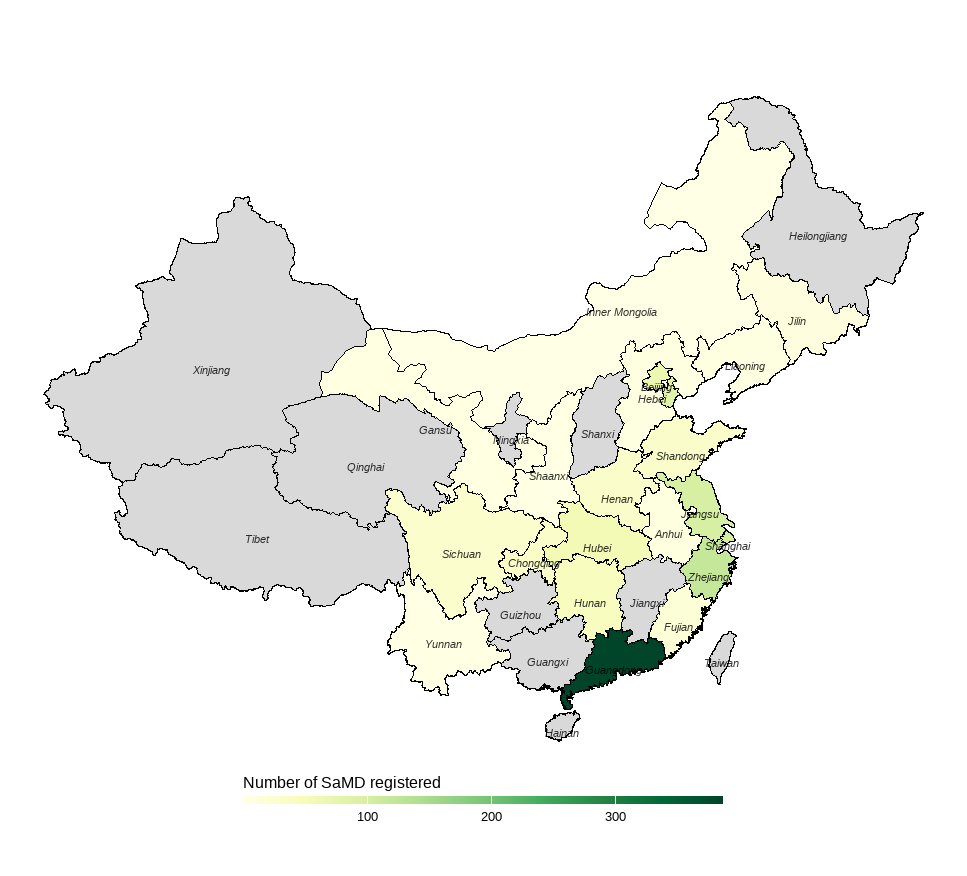}
    \caption{A geographic map of non-foreign medical device software are registration 2020--2024 (\emph{n} = 2,149). The areas include mainland provinces as well as the indication of ``Xu'' (the label used on the map for the Special Administration Regions).}
    \label{fig:map}
\end{figure}

In the filtered device list, we labeled each device register city and province, annotation for AI, non-AI, whether SaMD and SiMD and put into China map as shown in Figure \ref{fig:map}, we can see in each province software registration amount in different province of china map.

The analysis of medical device software (MDSW) reveals distinct patterns in the distribution and classification of Software as a Medical Device (SaMD) and Software in a Medical Device (SiMD). The dataset indicates that SaMD is predominantly domestic, with the majority classified under Class II, suggesting that most standalone software applications in the medical field are moderate-risk solutions rather than high-risk interventions. This aligns with regulatory expectations, as standalone software is often used for decision support, diagnostics, or treatment recommendations, rather than direct intervention in critical medical procedures. In contrast, SiMD is more evenly distributed across domestic and imported categories, with a higher proportion classified as Class III, indicating its integration into high-risk medical equipment.

In our analysis, AI-enabled devices were predominantly identified within standalone software (SaMD), with very few detected among software integrated into hardware (SiMD). This outcome is not due to a limitation of the search strategy, which was carefully designed, transparently documented, and reproducible. Rather, it reflects the fact that many SiMD product entries in the UDI database lack explicit references to AI-related terms such as “machine learning” or “intelligent analysis.” For example, well-known devices known to incorporate AI (e.g., deep learning reconstruction in imaging equipment) do not contain any such keywords in their official product descriptions. As such, even rigorous search methods are constrained by what is documented in the database. This finding highlights an important reality: the absence of identifiable AI in certain categories is not necessarily due to analytical shortcomings, but to how information is entered and structured in the registry. Future regulatory database enrichment efforts could help improve visibility and traceability of AI functionalities.

For AI devices, the majority of registrations are for domestic Class III devices, which constitute approximately 70\% of the total. This figure includes manufacturers both within and outside of Beijing, as per regulatory requirements mandating that all Class III devices must be registered in Beijing (national level). For Class II devices, the numbers decline progressively, with Guangdong and Hubei each accounting for around 8\% of the total registrations. Additional contributions come from Zhejiang (5\%), Tianjin (3\%), and Jiangsu (2\%), with Shanxi and Sichuan provinces each representing 1\% of the total registrations. 


\subsection{Artificial intelligence devices and algorithms in Medical software}
The authors cross-checked and validated the machine-filtered information, followed by a manual review to ensure accuracy. This process led to the development of a comprehensive database, which includes 43 AI-enabled devices, as shown in Figure \ref{fig:alluvial}. A detailed overview of these devices and their associated algorithms is provided in Appendix 2.

\begin{figure}[!t]
    \centering
    \includegraphics[width=\textwidth]{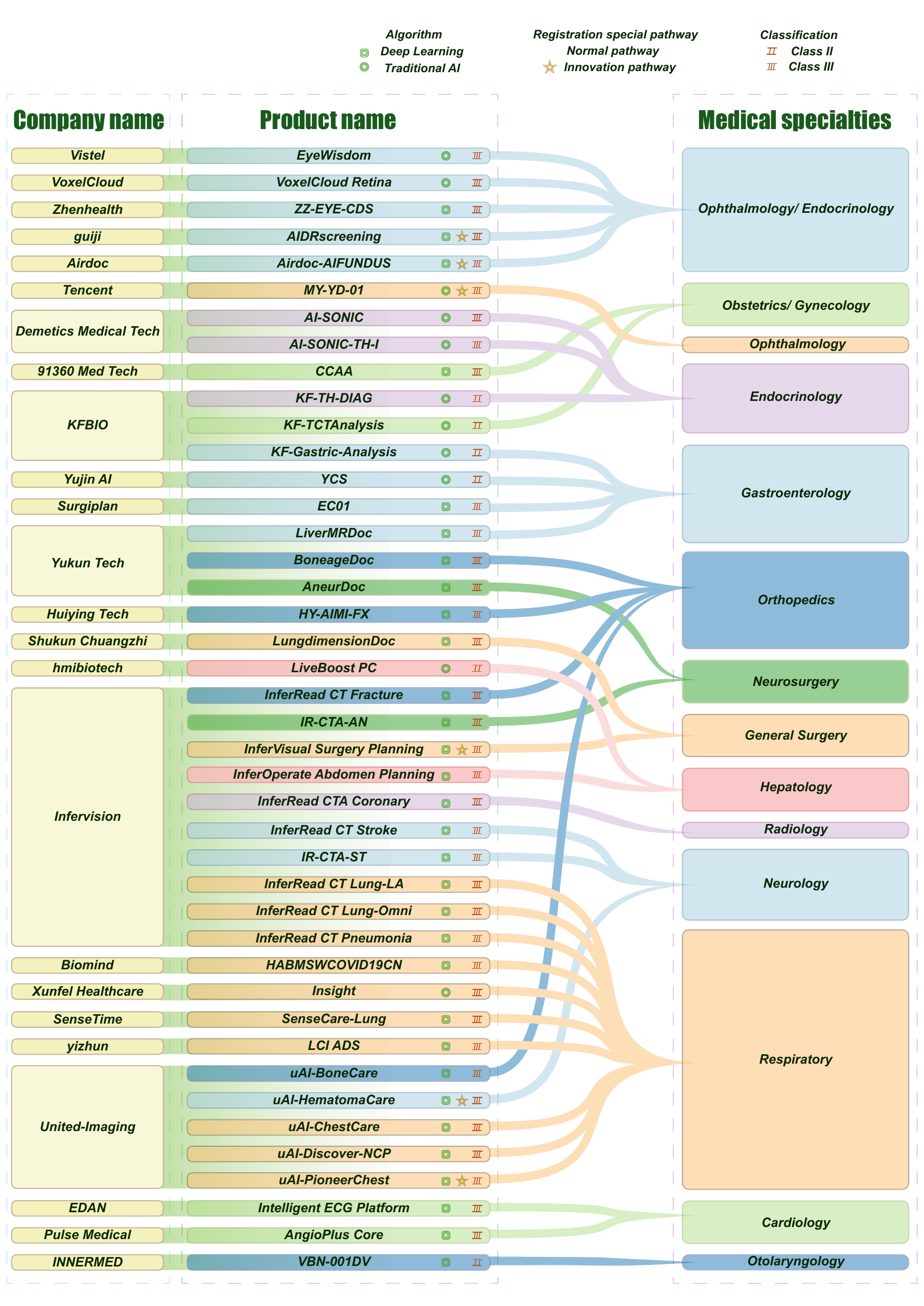}
    \caption{An alluvial diagram illustrating the types of 43 AI medical devices and the risk classifications along with the approval pathways within the NMPA regulatory framework}
    \label{fig:alluvial}
\end{figure}

Our approach introduces greater granularity by incorporating the specific anatomical or disease-related focus of AI-enabled medical devices. This distinction is valuable for both manufacturers and patients, as it provides clearer insights into the availability of advanced technologies for specific medical conditions. A medical specialty refers to a distinct branch of medical practice that focuses on a specific group of diseases, organ systems, diagnostic techniques, or therapeutic interventions \cite{torpy2007medical}. Specialties are often defined by the type of medical expertise required, the technology employed, and the patient population served. In the context of AI-enabled medical devices, specialty classification can be based on the primary body system or disease area targeted by the device, rather than solely on its function \cite{weisz2003emergence}.

The top three medical specialities that use AI-enabled medical devices are respiratory (20.5\%), Ophthalmology/Endocrinology (12.8\%), and orthopedics (10.3\%). Six AI-enabled medical devices (6 out of 43) have successfully utilized the innovation fast track pathway, which is eligible for underscoring the rapid advancement and regulatory approval of cutting-edge technologies in critical medical fields. All six devices are classified as Class III. All devices were registered after 2020, with the majority being registered in 2023 and 2024. This trend indicates that device registration activities are aligned with the regulations outlined in Table \ref{tab:regulations}.

\subsection{Exploratory Observations: Algorithm Types and Classification}

\begin{figure}[t!]
    \centering
\includegraphics[scale=0.6]{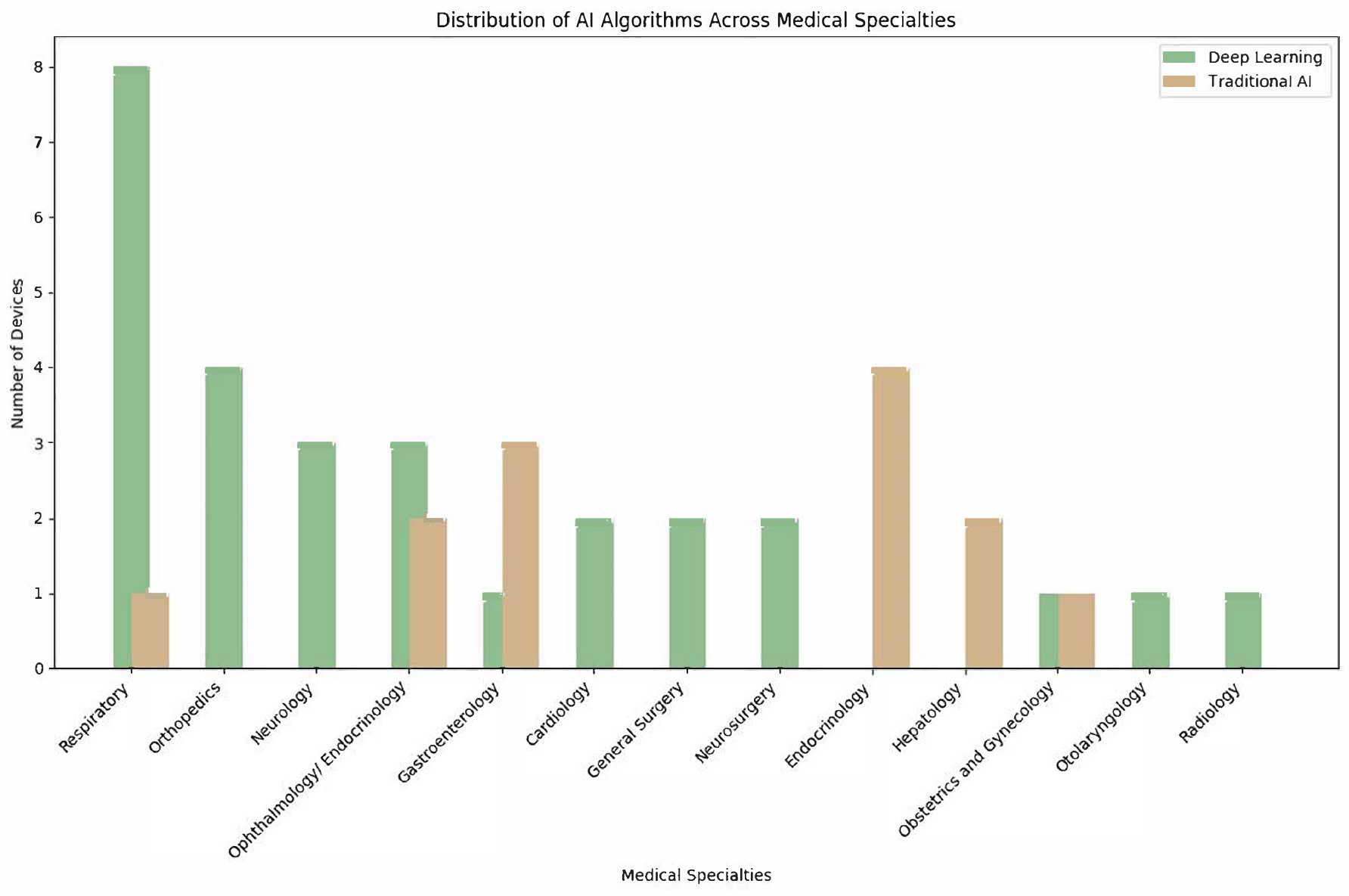}
\caption{Distribution of AIMD on different medical specialty and algorithm}
    \label{fig:deep}
\end{figure}

We examine descriptive trends across devices labeled as deep learning or traditional AI to identify any emerging differences in their regulatory pathways. These analyses are intended to generate hypotheses for future study rather than confirm causality. We reviewed each device in detail to label their underlying technology. Among the 43 AI-enabled medical devices (AIMD), we found that 32 utilize deep learning (DL), while 11 use traditional AI, as shown in Figure \ref{fig:deep}.

While our study was not designed to formally test the relationship between algorithm type and regulatory classification, this trend may warrant further investigation. We found that deep learning (DL) is applied across a variety of medical specialties, including respiratory imaging, orthopedics, neurology, and the combined fields of ophthalmology and endocrinology, as well as in cardiology, general surgery, and neurosurgery. In contrast, fields such as gastroenterology, obstetrics and gynecology, and otolaryngology tend to rely more on traditional AI techniques. This distribution raises important questions about why certain AI technologies are adopted in specific medical specialities.



Medical imaging techniques vary by machine and medical application. CT provides detailed cross-sectional views of bones, tissues, and blood vessels\cite{van2024impact}. MRI captures multi-planar images without radiation, ideal for diagnosing brain, spine, and musculoskeletal conditions. Ultrasound, using sound waves, is suited for real-time imaging of soft tissues and is portable for bedside use. Endoscopy allows visualization of internal cavities, enabling real-time diagnostics and interventions. Optical imaging, like fundus photography and OCT, is essential in ophthalmology for diagnosing retinal conditions\cite{greco2023imaging}.


\subsubsection{Deep learning and traditional AI algorithm analysis}
Traditional AI, encompasses a wide range of AI techniques and methods, including traditional machine learning algorithms (like decision trees, support vector machines), rule-based systems, and simple statistical models\cite{dinesh2024medical}. These techniques are applied to tasks where structured data, such as numerical or categorical data, can be analyzed to classify, predict, or detect patterns. Deep Learning, on the other hand, is a subset of machine learning that focuses on neural networks with multiple layers (also known as deep neural networks). Inspired by the human brain, these networks consist of interconnected layers of artificial neurons that learn to recognize features from raw data inputs through hierarchical abstraction\cite{zhou2023deep}. Deep learning models excel in processing unstructured data, such as images, audio, and text, due to their ability to learn hierarchical representations of data features. Deep learning, with its ability to process large, complex, and unstructured datasets, is particularly suited for specialities including ophthalmology, and neurology, where medical imaging are complex.

\subsubsection{Analysis of AI algorithm in certain medical specialty}
We extract medical specialties information automatically by identifying the disease location from the generic names. According to the "Rules for Generic Names of Medical Devices" (Order No. 19) \cite{CFDA2015}, characteristic terms describing specific attributes—such as the location of the disease where the software is used—are mandated to be included in the generic name. For example, a device named "Lung Nodule CT Image Auxiliary Detection Software" is classified under respiratory medicine, while "Chronic Glaucoma-like Optic Neuropathy Fundus Image Auxiliary Diagnosis Software" falls under ophthalmology.

We use this data-driven approach to achieve finer-grained classification of medical specialties compared to methods that categorize devices solely based on the underlying imaging or sensing technologies. For example, previous approaches often label most software that utilizes CT scans as radiology, regardless of the actual disease or organ system addressed, which limits downstream analyses of clinical applications \cite{benjamens2020state}.

From result, we found that deep learning (DL) is widely used in specialties like respiratory, orthopedics, and neurology. CT scans, with their high resolution and structured data, are ideal for convolutional neural networks (CNNs) \cite{iqbal2023analyses}. CNNs excel at feature extraction and segmentation, especially in 3D data, making them effective for detecting fractures, tumors, and lung abnormalities in chest, abdomen, and brain imaging.

In respiratory, DL is applied to CT scans and X-rays for tasks like lung cancer detection and pneumonia diagnosis \cite{thanoon2023review}, using supervised learning with labeled images. Orthopedics benefits from DL's ability to analyze X-rays, CTs, and MRIs for bone fractures and musculoskeletal conditions \cite{bousson2023application}. In neurology, DL analyzes brain MRIs to identify tumors, strokes, and Alzheimer's, using supervised learning with labeled data \cite{chaki2023deep}. In gastroenterology, DL is applied to endoscopic images to detect polyps, while traditional AI handles structured data like blood tests \cite{takahashi2023artificial}. Hepatology and endocrinology similarly rely on traditional AI for analyzing structured clinical data, such as liver function tests and hormone levels, aiding in disease monitoring and treatment planning \cite{schattenberg2023artificial,van2023thyroid}.

\begin{landscape}
\begin{figure}[ht]
    \centering
    \includegraphics[scale=0.8]{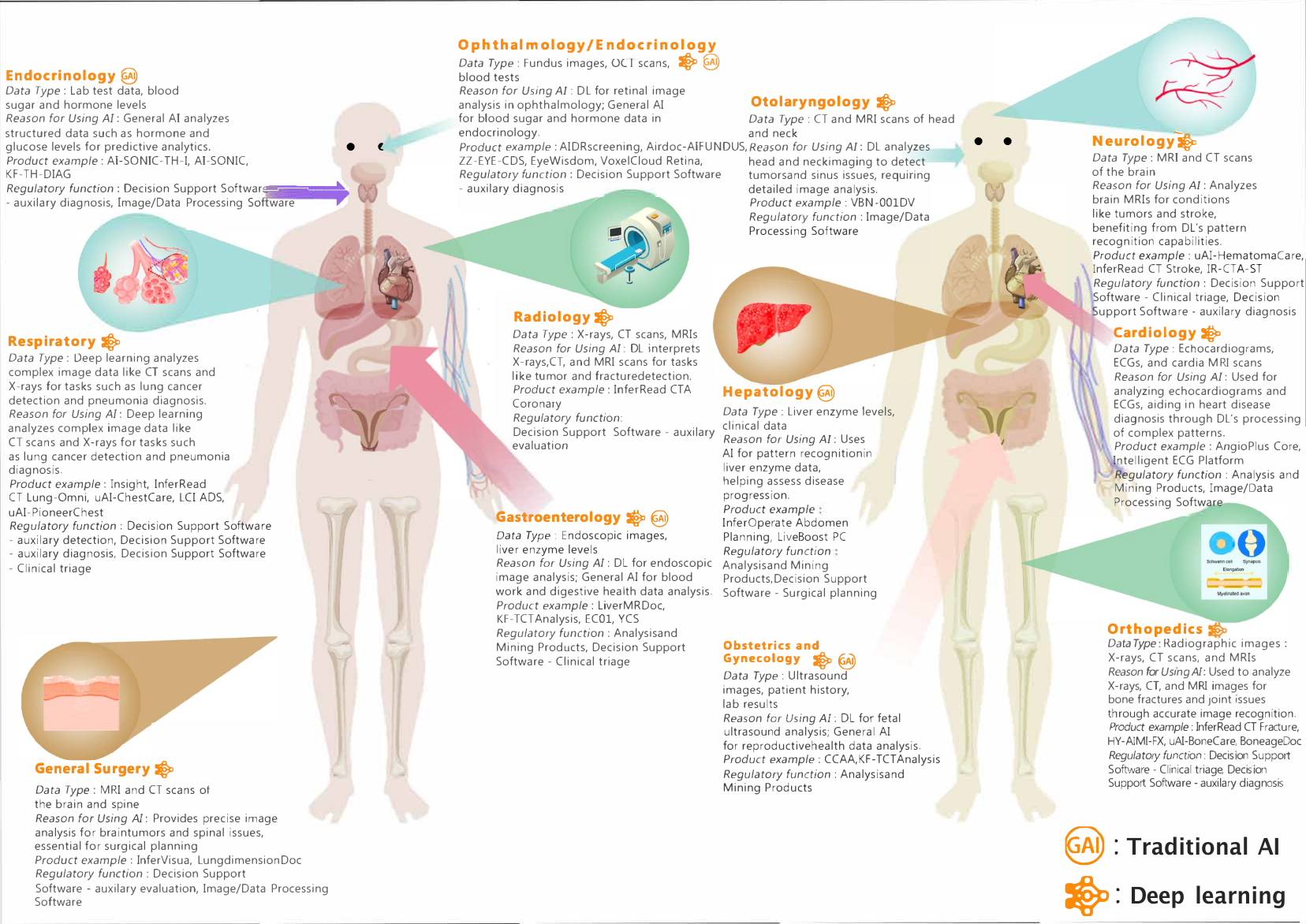}
    \caption{Applications of AI in Different Medical Specialties}
    \label{fig:b}
\end{figure}
\end{landscape}

Figure \ref{fig:b} presents a visual representation using a human body diagram, with key medical specialties highlighted and labeled to indicate the specific AI methods and technologies used within each domain. The diagram places these technologies directly near the anatomical regions they impact, offering a more intuitive understanding for readers.

\subsubsection{Analysis of AI devices}

From a functional standpoint, NMPA classifies AI-integrated software into specific functional categories, as outlined in Table \ref{tab:device_categories_count}. Of the 43 devices reviewed, 5 devices fall under the category of Image/Data Processing Software. For instance, INNERMED’s VBN-001DV device falls into this category and is classified as Class II, despite incorporating deep learning components. This classification indicates that when the AI algorithm primarily enhances image quality or focuses on segmentation, the NMPA may assign a lower risk classification, reflecting the specific regulatory approach.

30 of the devices fall under the category of Decision Support Software, which is primarily designed to assist medical decision-making.  This category can be further classified into subtypes based on functionality: auxiliary detection, auxiliary diagnosis, clinical triage, auxiliary evaluation, and surgical planning. For example, the “CT image-assisted triage software for intracranial hemorrhage” helps diagnose intracranial hemorrhage and supports triage decisions. The Insight, produced by Xunfel Healthcare, is one example. Although it does not utilize deep learning, it aids physicians by providing decision support for lung nodule detection in CT images, and the NMPA has classified it as Class III due to the function. Another device, the AI-SONIC-TH-I by Demetics Medical Tech, offers diagnostic support for thyroid nodule assessment using ultrasound images. Infervision’s InferVisual Surgery Planning device provides surgical planning support, including the display, processing, measurement, and analysis of chest CT images. It can automatically identify pulmonary nodules of 4mm and larger, showing their spatial relationship to key anatomical structures—information that thoracic surgeons can use in planning lung surgeries. This device has been granted classification under the innovation pathway, reflecting its advanced functionality in surgical planning support.

5 of these devices are categorized as Analysis and data Mining Products, which are primarily utilized for data analysis and mining. These devices normally process structured data through AI technology to help identify potential risk factors or predict disease progression. The "chronic liver disease risk calculation software" is used to assess the risk of chronic liver disease, supporting early clinical intervention.


The fast-track pathway designation has been awarded to several devices, including uAI-HematomaCare and uAI-PioneerChest from United-Imaging in the areas of neurology and respiratory, respectively; AIDRscreening by Guiji in the field of ophthalmology/endocrinology; MY-YD-01 from Tencent, also in ophthalmology; InferVisual Surgery Planning by Infervision for general surgery; and Airdoc-AIFUNDUS from Airdoc, again in ophthalmology/endocrinology. This trend may underscore the growing importance of ophthalmological AI technologies in China and reflects a global market demand for innovative ophthalmic solutions.

\begin{table}[h!]
\centering
\caption{Divide AIMD by software function from NMPA regulation}
\begin{tabular}{|p{4cm}|p{8cm}|p{2cm}|}
\hline
\textbf{Category} & \textbf{Description} & \textbf{Number of Devices} \\ \hline
\textbf{Decision Support Software} & Devices used for assisting medical decision-making that apply an AI algorithm, such as software for lesion identification and drug dosage calculation. & 30 \\ \hline
\quad - auxilary detection & Assists in detecting medical conditions, often using imaging data or other diagnostic inputs. & 1 \\ \hline
\quad - auxilary diagnosis & Supports diagnosis processes by providing probabilistic assessments or condition classification. & 16 \\ \hline
\quad - Clinical triage & Aids in triaging patients based on severity, prioritizing care delivery. & 10 \\ \hline
\quad - auxilary evaluation & Offers evaluation support, such as predicting treatment outcomes or assessing patient risk factors. & 1 \\ \hline
\quad - Surgical planning & Provides guidance and planning for surgical procedures through image analysis or patient-specific modeling. & 2 \\ \hline
\textbf{Image/Data Processing Software} & Devices used for medical imaging and data processing using AI algorithms, enabling tasks like image segmentation and fusion. & 5 \\ \hline
\textbf{Analysis and Data mining Products} & Devices used to analyze and mine medical-related data, applicable in fields like drug development, medical research, and hospital information management. & 5 \\ \hline
\textbf{Medical Assistant Products} & Based on electronic health records or patient information, these devices use NLP and other technologies to make logical inferences about the patient’s condition. & - \\ \hline
\end{tabular}
\label{tab:device_categories_count}
\end{table}




\begin{table}[t]
    \centering
        \caption{Foreign companies with software devices.}
\begin{tabular}{lccl}
\toprule
Company & Country & Classification & medical specialty\\
\midrule
Simemens Healthcare GmbH& USA & III  & Radiology \\ 
Nucletron B.V. & Netherlands & III  & Radiology \\ 
Vavian Medical System & USA & III & Radiology\\ 

MIM Software & USA & III  & Radiology \\ 
\bottomrule
\end{tabular}
    \label{tab:import}
\end{table}

\subsection{Observed Differences in International Regulatory Classification}
Table \ref{tab:import} presents AI-enabled medical devices (AIMDs) imported into China, the majority of which originate from the United States and are classified as Class III, typically under the domain of radiology and regulated centrally by the National Office in Beijing.

During our analysis, we noted several examples where the same AI-enabled device appeared to receive different classification levels across jurisdictions. For instance, the MIM Software system (Chinese license number National20153211878) is classified as Class III in China according to product code system (21-01-01), but is listed as Class II in the United States under 510(k) number K103576~\cite{fda_510k_database_medical_nodate}. These differences may reflect divergent interpretations of clinical risk or variation in regulatory frameworks, rather than inconsistencies in classification quality.

While our study does not aim to formally compare regulatory systems, these observations suggest that further research into regulatory divergence—especially for novel or AI-based medical technologies—could be valuable. The U.S. FDA permits Class II devices to be cleared via the 510(k) pathway by demonstrating substantial equivalence to existing approved devices. In contrast, NMPA frequently classifies decision-support AI systems as Class III and often requires clinical evaluations or trials, even for devices that are considered moderate-risk in other jurisdictions.

To offer additional context, we compiled Table \ref{tab:regulatory_comparison}, which outlines selected features of the FDA and NMPA regulatory frameworks. While the classification schemes are both risk-based and nominally similar, their implementation appears to vary in practice—particularly in the treatment of AI-enabled software as a medical device (SaMD).

A few illustrative cases underscore this variation: The Aidoc AI system for CT stroke detection is cleared as Class II via the 510(k) pathway in the U.S.~\cite{Aidoc2024}, but registered as Class III in China; Viz.ai’s stroke triage system, cleared under the De Novo pathway by the FDA~\cite{Squire2018}, is also classified as Class III by the NMPA; In contrast, Zebra Medical Vision’s radiology solutions are consistently treated as Class II in both jurisdictions~\cite{FDA_Zebra2020}; For high-risk applications such as HeartFlow’s AI-based CT analysis for coronary artery disease, both agencies assign a Class III designation~\cite{HeartFlow2024}.

These examples point to possible jurisdictional variation in perceived clinical risk and classification system. Many countries have faced challenges in consistently classifying AI-enabled medical devices~\cite{fink2023comparison}, ~\cite{gerke2020ethical}. These discrepancies in classification often reflect deeper differences in legal accountability and regulatory oversight. For instance, in the United States, liability for AI-assisted decisions is typically assessed through a tort-based framework, focusing on whether the standard of care was met. In contrast, other jurisdictions may apply stricter pre-market classification rules, driven by concerns over technological opacity or potential clinical risks~\cite{price2019liability}.

\begin{table}[h!]
\centering
\caption{Comparison of Medical Device Regulatory Frameworks: FDA  vs. NMPA}
\label{tab:regulatory_comparison}
\begin{tabular}{|p{4cm}|p{5cm}|p{5cm}|}
\hline
\textbf{Regulatory Factor} & \textbf{FDA (United States)} & \textbf{NMPA (China)} \\ \hline

\textbf{Regulatory Framework} & 
Medical devices are classified under 21 CFR Parts 800-898 & 
Classified based on the Medical Device Classification Catalogue  \\ \hline

\textbf{Classification System} & 
Three classes: I (low risk), II (moderate risk), III (high risk) & 
Three classes: I (low risk), II (moderate risk), III (high risk) \\ \hline

\textbf{Risk-Based Approach} & 
Risk assessment follows intended use and risk to patient safety & 
Similar risk-based approach, often more stringent for AI-driven decision-support devices \\ \hline

\textbf{Class I Devices} & 
Low risk: General controls (e.g., bandages, gloves) & 
Low risk: General controls (e.g., surgical masks) \\ \hline

\textbf{Class II Devices} & 
Moderate risk: General + special controls (e.g., X-ray systems) & 
Moderate risk: Requires technical review (e.g., diagnostic ultrasound systems) \\ \hline

\textbf{Class III Devices} & 
High risk: PMA required  & 
High risk: Technical review + clinical trials  \\ \hline

\textbf{Approval Pathways} & 
- \textbf{510(k)} for Class II (demonstrates substantial equivalence to a predicate device) \newline
- \textbf{De Novo} for novel moderate-risk devices \newline
- \textbf{PMA} (Pre-Market Approval) for Class III high-risk devices & 
- \textbf{Class II \& III} require clinical evidence \newline
- \textbf{Class III} demands full clinical trials unless exemptions apply \\ \hline

\textbf{Clinical Evidence Requirements} & 
- \textbf{Class I}: Typically exempt \newline
- \textbf{Class II}: Bench testing, software validation, predicate comparison \newline
- \textbf{Class III}: Clinical trial data, human factors validation, risk mitigation plans & 
- \textbf{Class I}: Exempt \newline
- \textbf{Class II}: Regulatory testing, clinical evaluation \newline
- \textbf{Class III}: Mandatory clinical trials unless explicitly waived \\ \hline

\textbf{SaMD Regulation} & 
Assessed per device classification and IMDRF SaMD framework & 
Specific guidance for AI/software under NMPA rules \\\hline

\textbf{Foreign Device Registration} & 
Requires approval in origin country or extensive documentation & 
Requires origin approval + local clinical trials (in some cases) \\ \hline

\textbf{UDI Implementation} & 
Mandatory UDI for all classes & 
Gradual UDI adoption; regional variations exist \\ \hline

\textbf{Harmonization Efforts} & 
Member of IMDRF (International Medical Device Regulators Forum), moving towards global alignment & 
Member of IMDRF, yet retains national requirements \\ \hline
\label{tab:diff}
\end{tabular}
\end{table}

\textbf{Disclaimer:} All observations in this section are descriptive and exploratory in nature. No inferential claims or normative judgments are made. These results are presented to support hypothesis generation and future research in regulatory science.

\section{Discussion}
\label{sec:discussion}
\subsection*{The automated method and limitation}
The automated search process demonstrated significant efficiency, particularly in processing a vast dataset of 4 million devices. Compared to previous studies \cite{benjamens2020state}, our method applies a structured filtering approach designed to improve reproducibility and consistency, with further validation to assess its accuracy across broader device categories. First, we systematically identified software-based devices and applied the 21-code rule to precisely extract Software as a Medical Device (SaMD)—a highly accurate step that is often overlooked by other methodologies without regulatory background author. This rule allowed us to confidently isolate the standalone SaMDs from the broader dataset. Subsequently, we further refined our search by implementing keyword-based filtering on the resulting medical software products, effectively pinpointing AI-enabled medical devices. This automated multi-step approach enabled us to reduce the dataset to a manageable and highly relevant subset while ensuring further rigorous selection process.

A key strength of this study is it is the first study in regulatory science about application of data science methods to map license numbers and automate the filtering of devices, offering a reusable process that contrasts with manual approaches or keyword-based methods used in other studies \cite{liu2024regulatory}, \cite{benjamens2020state}. using the code, reserachers could use their own keyword strategy and do the analysis at ease. While we do not claim that the current semi-quantitative automatic approach outperforms manual expert judgment, we believe it represents a necessary and bold step toward standardizing the classification process and generating reproducible insights. As regulatory data infrastructures evolve, combining quantitative tools with domain expertise may offer a promising path forward.

\subsection*{False positive results and regulatory term}
In the final filtering stage, after applying keyword-based identification, we initially identified 75 devices. However, upon manual review, 32 devices were excluded(a false-positive rate of approximately 43\%). The removal of these devices was not due to a fundamental flaw in our methodology but rather a result of the regulatory terminology “auxiliary diagnosis” and “auxiliary treatment,” which are explicitly defined within NMPA regulations as inclusion criteria for medical devices. AI-enabled devices frequently incorporate this terminology in their official naming conventions \cite{naming}. According to regulatory guidelines, AI is presumed to be involved in auxiliary functions but is not necessarily the primary functional component of the device. It is crucial to distinguish that not all devices labeled with auxiliary functions are genuinely AI-enabled. This relationship follows a logical structure: $A \subseteq B$, but not necessarily $A = B$—meaning that while all AI-enabled devices for auxiliary diagnosis or treatment are captured, not all devices labeled for auxiliary functions necessarily utilize AI. Consequently, the primary source of false positives in our analysis arises from this regulatory language rather than a limitation in the automated filtering approach. Researchers employing alternative methodologies that do not rely on these specific regulatory terms would likely not encounter this issue. For example, one device manually excluded from our dataset was a red light therapy system. Its product description contained the term “auxiliary treatment,” yet it did not incorporate any AI-based functionality. The description stated: “The red light therapy device consists of hardware and software components, typically including a light radiation source (such as LEDs), a control unit, and a support structure (which may include positioning components). It may also be equipped with light-guiding accessories. By utilizing red light wavelengths to irradiate gynecological areas (some devices may also include infrared wavelengths), it induces photochemical reactions and/or biological stimulation within human tissues, achieving an auxiliary treatment effect.”

\subsection*{Limitations from data side and Why AI is currently absent in SiMDs}

One key limitation of this study is that the dataset is derived from the UDI (Unique Device Identification) system, which does not provide a comprehensive representation of the entire medical device industry. The dataset is heavily influenced by how manufacturers choose to report their product information and how regulatory authorities define and enforce reporting requirements. This variability in data entry practices affects the completeness and accuracy of AI-related classifications within regulatory databases. Through this study, we aim to highlight these limitations, providing insights for industry professionals, researchers, and regulatory bodies to enhance AI medical device classification and reporting standards.

Due to the absence of standardized algorithm classification in the UDI database, our classification relies on product descriptions as proxies, which may not always align with technical reality. This limitation is an inherent feature of the regulatory reporting system and highlights the need for improved transparency and labeling standards.

Our investigation into medical devices registered within the UDI system—including GE Vivid ultrasound, Canon MRI and CT scanners, and Edwards Lifesciences cardiac surgical monitors — revealed challenges in aligning known AI functionalities with information available in regulatory descriptions, when repurposing the database for identifying AI-enabled devices. Canon’s MRI and CT scanners integrate Deep Learning Reconstruction (DLR) and AICE noise reduction algorithms, which enhance image quality and diagnostic accuracy \cite{Canon_AICE_DLR}. GE’s MRI systems feature AIR Recon DL, an AI-powered reconstruction technique designed to improve image sharpness while reducing scan times \cite{GE_AIR_Recon_DL}, GE Vivid ultrasound systems incorporate AI-based functionalities, including automated segmentation and measurement of left ventricular ejection fraction (LVEF), which enhance cardiac imaging and analysis \cite{GE_Vivid_Ultrasound}. Edwards Lifesciences' cardiac surgical monitors employ the Acumen Hypotension Prediction Index (HPI), which utilizes predictive analytics to forecast hemodynamic instability during surgery \cite{Edwards_HPI}. While these products are known to incorporate AI functionalities (e.g., deep learning-based reconstruction or segmentation), the corresponding UDI records do not consistently include explicit references to such terms. This reflects a limitation when repurposing the UDI database for secondary research tasks such as identifying AI-enabled features.

These findings highlight AI-enabled features are not systematically labeled in product descriptions, which presents a limitation when using the UDI database for secondary purposes such as identifying AI-enabled devices. As AI adoption in medical imaging and monitoring continues to grow, further refinement in regulatory reporting could improve transparency and facilitate AI-specific assessments within the UDI framework. This partially constitute why in this data analysis, AI is currently absent in SiMDs.

\subsection*{Descriptive Patterns of Algorithm Use in Medical Specialties}
The classification of devices into deep learning and traditional AI was based on the terminology present in the medical device database and supplemented by expert review.

Devices were classified as deep learning-based if their product descriptions explicitly mentioned terms such as: Deep Learning, CNN, RNN. Traditional AI Classification refer to Devices were classified into this category if they referenced. Artificial Intelligence or Machine Learning without specifying deep learning techniques, other Statistical and rule-based AI models. The classification was not directly provided by the database but was instead inferred from the textual descriptions and keywords found within the dataset.

From data, We observed that deep learning-based devices were more frequently categorized as Class III (high-risk) medical devices, particularly in imaging-related applications. Traditional AI devices, such as rule-based decision support systems, had a broader distribution across Class II and Class III categories. DL devices predominantly used for automated image analysis, segmentation, and real-time diagnostic assistance. Traditional AI devices were more frequently associated with predictive modeling, decision-support systems, and clinical workflow optimization.

From a clinical perspective, it remains an open question whether end-users—such as physicians—engage deeply with the technical architecture of AI-based tools. While regulatory agencies increasingly emphasize algorithmic explainability, it is unclear whether such transparency translates into meaningful clinical understanding or alters decision-making. Whether clinicians make a “second judgment” beyond the model’s recommendation remains a critical area for future interdisciplinary research involving HCI, AI transparency, and medical education.

\subsection*{Descriptive Trends: Deep Learning and Classification Level}

Our analysis of regulatory data reveals patterns in the classification of AI-enabled medical devices. While the assignment of Class II or Class III status is officially based on intended use and associated risk, we observe that certain types of AI algorithms, such as deep learning, appear more frequently among Class III devices. This may reflect their common use in higher-risk applications—including autonomous diagnosis, treatment planning, or real-time monitoring—though we emphasize that algorithm type is not the sole determinant of classification. Conversely, deep learning algorithms are also found in Class II devices, particularly where the application is lower-risk (e.g., workflow support or image enhancement). Similarly, some Class III devices do not use deep learning at all, indicating that traditional AI methods can still be associated with significant clinical risk when applied in complex or high-stakes scenarios.

We also observe that domestically manufactured Class III AI-enabled SaMDs outnumber imported ones, while the distribution of AI-enabled SiMDs appears more balanced. This observation suggests differences in market strategy or regulatory positioning, but further research would be needed to assess the causes. Moreover, our analysis highlights that device records in the UDI system often lack consistent labeling regarding AI usage. This may pose challenges for data-driven research and underscores the importance of transparent reporting in regulatory documentation. Finally, we find that traditional AI techniques remain prevalent in domains such as gastroenterology and endocrinology, where clinical decisions rely more heavily on structured data and established protocols. In contrast, deep learning appears to play a growing role in more complex diagnostic specialties. These patterns merit further exploration in future work.

Deep learning (DL) models are often criticized for their "black-box" nature, as their complex architectures and high-dimensional feature spaces can make it difficult to interpret how decisions are made. In contrast, traditional AI techniques—such as decision trees, rule-based systems, or logistic regression—typically offer greater transparency and interpretability. These differences are not merely technical; they have direct implications for patient safety, clinical accountability, and regulatory oversight. For instance, when a model's decision cannot be readily explained, it becomes more challenging for healthcare providers to validate its recommendations or justify its use in clinical practice \cite{wysocki2023assessing}. This opacity complicates legal attribution of responsibility in adverse events and often prompts regulators to impose stricter pre-market scrutiny or require post-market surveillance for DL-based systems. Conversely, the interpretability of traditional AI may facilitate clearer pathways for risk assessment and compliance.

The analyses of algorithm type and regulatory differences were exploratory in nature, and we do not draw generalizable conclusions. We acknowledge the limitations in data labeling, jurisdictional scope, and sample representativeness.

\subsection*{Interpretability Demands Across Clinical Specialties: Regulatory and Clinical Implications}

Among the 43 AI-enabled medical devices analyzed, 32 incorporate deep learning algorithms. Of these, 21 are designed primarily for diagnostic purposes, as detailed in Appendix 2.  Among the 32 deep learning-based devices identified, 21 were classified as diagnostic tools, primarily concentrated in radiology, ophthalmology, and respiratory medicine—fields that traditionally rely on visual pattern recognition and tolerate a certain degree of algorithmic opacity. Conversely, the 11 devices utilizing traditional AI (e.g., rule-based logic, statistical models) were more evenly distributed across specialties such as endocrinology, cardiology, gastroenterology, and hepatology. These fields are characterized by structured clinical decision pathways, where physicians demand transparent justifications for each computational recommendation.

Clinically, specialties such as endocrinology and cardiology often require high levels of traceability and causality, especially in scenarios involving chronic disease management, dosage adjustment, and life-sustaining interventions. In these cases, the acceptance of opaque models such as deep neural networks remains limited. Regulatory authorities, such as China’s NMPA, tend to classify AI systems deployed in such high-stakes clinical areas as Class III, regardless of whether deep learning or traditional AI is used. This reflects the elevated regulatory burden associated with low-interpretability systems in high-risk domains. Specialties like radiology, ophthalmology, and orthopedic imaging have shown a greater tolerance for black-box models, given that such tools function primarily as decision aids rather than autonomous decision-makers. Devices in these domains are more frequently classified as Class II, particularly when AI is used for pre-screening, image enhancement, or lesion detection with a physician-in-the-loop approach. Our findings suggest that interpretability requirements are not only a technical constraint but also a regulatory variable that influences classification, approval pathway, and post-market surveillance demands as shown in Table \ref{tab:interpretability}. 

These observations align with broader regulatory trends in the United States and European Union, where agencies have begun to tailor evidence requirements based on algorithm type and clinical application. For instance, the U.S. FDA’s Total Product Lifecycle (TPLC) framework incorporates transparency as a critical attribute in assessing AI/ML-enabled devices \cite{fda2024tplc}. Similarly, the EU Medical Device Regulation (MDR) places increased emphasis on algorithm accountability in AI systems with diagnostic intent \cite{european2022aihealthcare}. The findings suggest that the clinical demand for interpretability—and its translation into regulatory risk—is specialty-specific and context-sensitive. As AI tools become more complex and deeply integrated into healthcare workflows, developers must consider not only model accuracy but also the interpretability thresholds imposed by clinical end-users and regulators alike.

\begin{table}[htbp]
\centering
\caption{Interpretability Demands Across Clinical Domains}
\smallskip

\textbf{A. Clinical domains with high interpretability requirements} \\
\vspace{2pt}
\begin{minipage}{\textwidth}
\small
These domains typically involve high-stakes diagnostic or therapeutic decisions, elevated legal liability, and require physicians to clearly understand and trace system-generated outputs.
\end{minipage}

\begin{tabular}{|p{3.5cm}|p{5.5cm}|p{4.5cm}|}
\hline
\textbf{Clinical Domain / Scenario} & \textbf{Rationale} & \textbf{Example} \\
\hline
Endocrinology (e.g., diabetes management) & Requires precise control of physiological variables (e.g., glucose, insulin); physicians rely on rule-based models & Medication dosage adjustment software \\
\hline
Cardiology & High traceability required for arrhythmia detection and treatment plans & ECG analysis, CHADS\textsubscript{2} risk scoring systems \\
\hline
Emergency / ICU & Demands rapid responses with clear intervention justifications & Shock index scoring, hospital readmission prediction \\
\hline
Clinical pathway management & Follows guidelines and decision trees for treatment standardization & Insulin injection decision-support systems \\
\hline
Pathology (non-imaging) & Emphasizes structured classification based on clinical logic & Interpretative systems for pathology reports \\
\hline
\end{tabular}

\vspace{0.8cm}

\textbf{B. Clinical domains with lower interpretability requirements} \\
\vspace{2pt}
\begin{minipage}{\textwidth}
\small
These domains emphasize visual pattern recognition or large-scale empirical learning. Clinicians are generally receptive to AI-assisted outputs based on prior experience with diagnostic imaging tools.
\end{minipage}

\begin{tabular}{|p{3.5cm}|p{5.5cm}|p{4.5cm}|}
\hline
\textbf{Clinical Domain / Scenario} & \textbf{Rationale} & \textbf{Example} \\
\hline
Radiology & Complex imaging features; clinicians already accustomed to AI as assistive tools & AI-assisted CT / MRI lung nodule detection \\
\hline
Ophthalmology & Highly structured visual input (e.g., retina); facilitates standardized model training & Diabetic retinopathy screening algorithms \\
\hline
Dermatology & Distinctive visual features; AI outputs easily understood and accepted & Melanoma detection tools \\
\hline
Preoperative Surgical Planning & Used primarily for visualization; not directly responsible for clinical decision-making & 3D visualization of pulmonary nodules for surgical planning \\
\hline
Orthopedic Imaging & Structural changes (e.g., fractures) are easily quantifiable & Bone age estimation tools using radiographs \\
\hline
\end{tabular}
\label{tab:interpretability}
\end{table}

\subsection*{Fast-track approvals}

Among the six fast-track approvals in China, three focus on ophthalmology, reflecting the demand for AI technologies like Guiji's AIDR screening device for diabetic retinopathy. The global market for such diagnostic software is projected to grow steadily, indicating strong market acceptance in China, in line with the New Generation Artificial Intelligence Development Plan \cite{QYResearch2024}.

\section{Conclusions}

Benjamens et al.\ \cite{benjamens2020state} manually reviewed FDA records and identified 64 AI/ML-based medical devices, validating them through multiple sources. In contrast, Muehlematter et al.\ \cite{muehlematter2021approval} reported approximately 500 AI medical devices across U.S. and European regulatory databases, although they did not clearly document their filtering or validation procedures. Benjamens et al.\ also highlighted the imprecision of keyword searches and emphasized the need for a dedicated AI labeling system to enable more effective device tracking.

To fully leverage data science for regulatory insights, a standardized labeling or coding system for AI functionalities is essential. While China launched an AI medical device registration platform in 2019~\cite{pla}, no comprehensive, publicly accessible AI-focused medical device database currently exists. In contrast, the U.S. FDA has published a centralized and regularly updated list of approved AI/ML-enabled medical devices, which provides a more transparent starting point for research~\cite{FDA_AI_ML_Devices}.

One limitation of our study stems not from the coding process itself but from the structure of the NMPA UDI database. Devices are submitted and published in batches, and some products may be missing from the dataset due to delays or incomplete reporting. This likely accounts for discrepancies between our findings and those of previous studies~\cite{liu2024regulatory}. While the UDI dataset remains the most comprehensive registry of medical devices in China, our findings underscore the need for a more complete and AI-specific regulatory infrastructure~\cite{benjamens2020state}.

As AI medical devices raise growing concerns around algorithmic transparency, bias, cybersecurity, and ethical use~\cite{dalton2020ethics}, the need for robust surveillance through dedicated regulatory databases becomes increasingly urgent. Questions of data privacy, fairness, and accountability—as well as the impact of black-box decision-making on the doctor-patient relationship—remain unresolved~\cite{stern2019cybersecurity}. Public excitement around general-purpose AI tools like ChatGPT~\cite{kung2023performance} has only intensified attention on AI’s role in healthcare. In this context, the rapid proliferation of AI-enabled devices demands more coordinated regulatory oversight and public communication.

Our primary contribution lies in offering a scalable framework for identifying and classifying AI-related devices using real-world regulatory data. Rather than aiming to produce a definitive list of all AI devices in China, we demonstrate how structured filtering methods can surface meaningful patterns—and simultaneously reveal systemic limitations in the current data infrastructure. This study represents an early but necessary step toward data-driven regulatory analysis at scale.

Our findings also suggest that even well-known AI-enabled devices may be missed under current conditions, simply because their product descriptions lack standardized terminology. This is not a flaw of the method, but a symptom of broader limitations in how AI capabilities are currently documented in regulatory systems. We therefore advocate for the enhancement of regulatory databases through dedicated AI labeling fields, consistent terminology, and clearer guidance for manufacturers.

Such improvements would not only enhance reproducibility and transparency for researchers but also support more rigorous oversight, safer deployment, and more informed policy. We hope this work will encourage collaboration among regulators, industry leaders, and researchers to build a more robust regulatory ecosystem—one capable of keeping pace with rapid advances in AI medical technology.

\section*{Competing interests}
The authors declare no competing interests. 

\section*{DATA AVAILABILITY}
The scripts used to carry out the analysis and the resultant data that support the findings in this study are available on: https://github.com/Oxford-NIL/NMPA-analysis



\section*{Declaration of generative AI and AI-assisted technologies in the writing process}

During the preparation of this work the author(s) used ChatGPT 4o in order to edit sentence grammar. After using this tool/service, the author(s) reviewed and edited the content as needed and take(s) full responsibility for the content of the publication.

\bibliographystyle{sn-vancouver}
\bibliography{medical}

\end{document}